\documentclass[letterpaper]{article} % DO NOT CHANGE THIS
\usepackage{aaai2026}
\usepackage{booktabs} 
\usepackage{times}  % DO NOT CHANGE THIS
\usepackage{helvet}  % DO NOT CHANGE THIS
\usepackage{courier}  % DO NOT CHANGE THIS
\usepackage[hyphens]{url}  % DO NOT CHANGE THIS
\usepackage{graphicx} % DO NOT CHANGE THIS
\usepackage{natbib}  % DO NOT CHANGE THIS AND DO NOT ADD ANY OPTIONS TO IT
\usepackage{caption} % DO NOT CHANGE THIS AND DO NOT ADD ANY OPTIONS TO IT
\usepackage{makecell}
\usepackage[table]{xcolor}

\usepackage{algorithm}
\usepackage{pifont}
\usepackage{multirow}
\usepackage{amsmath}
\usepackage{amssymb}
\usepackage{amsthm}
\usepackage{tabularx}
\newtheorem{definition}{Definition}
\newtheorem{theorem}{Theorem}
\usepackage{algpseudocode}
\usepackage{xcolor}
\newcommand{\sectionname}{Section}
\DeclareMathOperator*{\argmax}{argmax}

\usepackage{newfloat}
\usepackage{listings}
\DeclareCaptionStyle{ruled}{labelfont=normalfont,labelsep=colon,strut=off} % DO NOT CHANGE THIS
\floatstyle{ruled}
\newfloat{listing}{tb}{lst}{}
\floatname{listing}{Listing}
\title{Counterfactual Explainable AI (XAI) Method for Deep Learning-Based Multivariate Time Series Classification}
\author {
    Alan G. Paredes Cetina\textsuperscript{\rm 1},
    Kaouther Benguessoum\textsuperscript{\rm 1},
    Raoni Lourenço\textsuperscript{\rm 1},
    Sylvain Kubler\textsuperscript{\rm 1},
}
\affiliations {
    \textsuperscript{\rm 1}SnT, University of Luxembourg\\
    alan.paredes@uni.lu, kaouther.benguessoum@uni.lu, raoni.lourenco@uni.lu, sylvain.kubler@uni.lu
}

\begin{document}

\maketitle

\begin{abstract}
Recent advances in deep learning have improved multivariate time series (MTS) classification and regression by capturing complex patterns, but their lack of transparency hinders decision-making. Explainable AI (XAI) methods offer partial insights, yet often fall short of conveying the full decision space. Counterfactual Explanations (CE) provide a promising alternative, but current approaches typically prioritize either accuracy, proximity or sparsity -- \emph{rarely all} -- limiting their practical value. To address this, we propose \textsc{confetti}, a novel multi-objective CE method for MTS. \textsc{confetti} identifies key MTS subsequences, locates a counterfactual target, and optimally modifies the time series to balance prediction confidence, proximity and sparsity. This method provides actionable insights with minimal changes, improving interpretability, and decision support. \textsc{confetti} is evaluated on seven MTS datasets from the UEA archive, demonstrating its effectiveness in various domains. \textsc{confetti} consistently outperforms state-of-the-art CE methods in its optimization objectives, and in six other metrics from the literature, achieving $\geq10\%$ higher confidence while improving sparsity in $\geq40\%$. 
\end{abstract}

  \begin{links}
      \link{Code}{https://github.com/serval-uni-lu/confetti}
% %     \link{Datasets}{https://aaai.org/example/datasets}
% %     \link{Extended version}{https://aaai.org/example/extended-version}
 \end{links}

\section{Introduction}
%Multivariate time series (MTS) prediction problems are prevalent across various domains, including manufacturing, healthcare, finance, and economics \cite{wei2018multivariate}. Recently, there has been growing interest in leveraging 
Deep Learning (DL) models, such as Convolutional Neural Networks (CNNs) \cite{7966039}, Recurrent Neural Networks (RNNs) \cite{liu2020dstp}, and Transformer-based models \cite{schäfer2018multivariatetimeseriesclassification}, have recently shown strong performance in multivariate time series (MTS) tasks \cite{ismail2019deep,khan2023end}. However, their lack of transparency makes it hard for decision makers to interpret predictions \cite{ruiz2021great}. Explainable AI (XAI) methods help decision makers understand the rationale behind DL model predictions \cite{adadi2018peeking}. Even high-quality explanations may not reveal the full decision space. For example, an AI system might advise a broker to sell a stock due to recent volatility, and XAI could highlight the spike as the key factor. Yet the broker might not realize that a slight drop in volatility could have changed the recommendation. Counterfactual Explanations (CE), a subset of XAI, help fill this gap by showing how small input changes can alter predictions, thus revealing critical decision factors \cite{spinnato2022explaining}.

 %For MTS, CE could show how adjustments in specific time-series variables affect the prediction, offering more actionable insights. Current efforts to generate CE for MTS use several approaches to improve interpretability, including shapelet-based techniques \cite{bahri2022shapelet}, attention mechanisms \cite{li2023attention}, and hybrid methods that combine instance-based and rule-based explanations \cite{spinnato2023understanding}. A key limitation of these techniques is that they tend to focus on a single objective: either (i) maximizing prediction confidence: ensuring that the CEs are accurate but potentially requiring significant changes to the original time series; or (ii) focusing on sparsity: minimizing the number of changes in the time series, which may reduce the accuracy and relevance of the explanations. This narrow focus limits the overall effectiveness and usability of CE, as both prediction confidence and sparsity are essential for generating meaningful, actionable, and understandable insights.
State-of-the-art CE methods for MTS mostly rely on shapelet-based techniques \cite{bahri2022shapelet}, attention mechanisms \cite{li2023attention}, or hybrid instance- and rule-based approaches \cite{spinnato2023understanding}. However, they often optimize a single objective, either (i) maximizing \emph{prediction confidence}, which may require large changes in the original time series, (ii) promoting \emph{sparsity}, which can reduce explanation accuracy, or (iii) improving \emph{proximity}, which focuses on minimizing the distance from the original instance but may lead to out‑of‑distribution instances. This narrow focus limits the usefulness of CE, as all of these objectives are important to be jointly considered to generate actionable and understandable insights.

To address the lack of multi-objective CE methods for MTS classification, we introduce a novel method called \textsc{confetti} (COuNterFactual Explanations for mulTivariate Time serIes). It employs a four-step approach: it first locates the closest instance from the opposite class as the counterfactual target, then uses Class Activation Maps (CAMs) \cite{zhou2016learning} to identify the most influential subsequences. Next, it substitutes values from the target into the original series to create an initial CE, and finally optimizes this CE to balance prediction confidence, proximity, and sparsity, while ensuring plausibility, and validity by design.

In summary, the contributions of the paper are:
\begin{itemize}
    \item A novel multi-objective CE method for MTS that optimizes prediction confidence, proximity, and sparsity, while  ensuring plausibility and validity by design;
    \item A comprehensive benchmark against CoMTE, SETS, and TSEvo CE methods conducted on seven datasets from the UEA archive using two model architectures;
    \item An ablation study of the initial stage showing how the search for most influential subsequence through feature weights affects performance.
    \item A sensitivity analysis of our method parameters, examining how variations in their values affect performance;
\end{itemize}

\sectionname~\ref{sec:LR} provides a review of the CE literature for MTS. \sectionname~\ref{sec:problem} introduces the concepts used in our approach and formally define the problem. \sectionname~\ref{sec:CONFETTI} details the workings of \textsc{confetti}. \sectionname~\ref{sec:Exp} presents the experimental setup and results. \sectionname~\ref{sec:concl} summarizes our contributions.

\begin{figure*}[t]
    \centering
    \includegraphics[width=\textwidth]{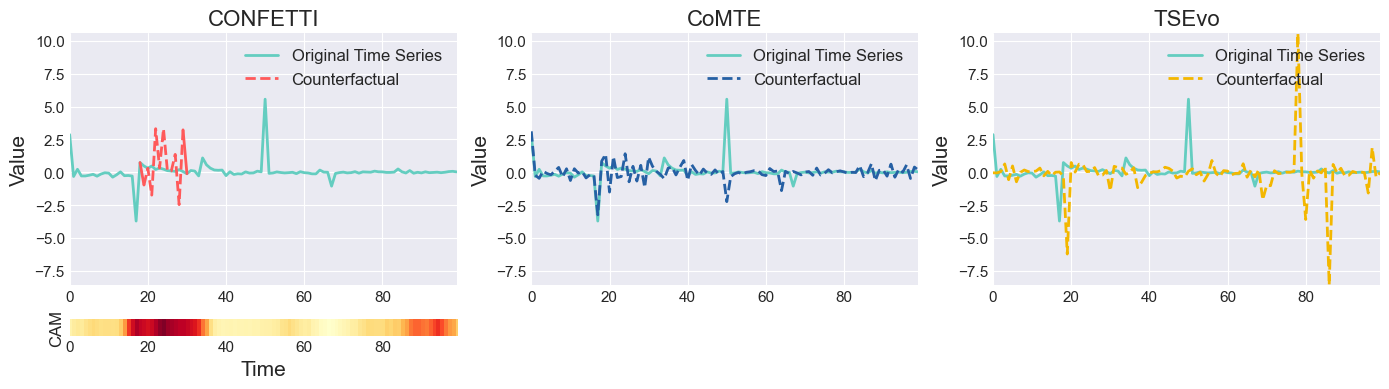}
    \caption{Counterfactual explanations generated by i) \textsc{Confetti}; ii) CoMTE; and iii) TsEVO for a single dimension of a multivariate time series. Below \textsc{Confetti} example is the Class Activation Map of the Nearest Unlike Neighbour as a heatmap, showing in red the most relevant timesteps.}
    \label{fig:example}
\end{figure*}

\section{Related Work} \label{sec:LR}

%\paragraph{State-of-the-art CE methods} 
CE are expected to satisfy several desiderata outlined in the XAI literature \cite{guidotti2024counterfactual}: (i) \textit{Validity}: the CE results in a different predicted class compared to the original instance; (ii) \textit{Confidence}: the predicted probability associated with the target class should be sufficiently high, providing a measure of how strongly the model supports the CE outcome; (iii) \textit{Sparsity}: the number of changes relative to the original instance is minimal; (iv) \textit{Proximity}: the magnitude of these changes is as small as possible; (v) \textit{Plausibility}: the CE should have feature values that lie within the distribution of a reference population, avoiding unrealistic values.

CoMTE~\cite{ates2021counterfactual} was the first CE method for MTS, which substitutes one channel of a MTS instance at a time with the corresponding channel of the nearest unlike neighbor (NUN). SETS \cite{bahri2022shapelet} uses the Shapelet Transformer to identify the most relevant subsequences per class (shapelets). Then, it substitutes the identified shapelets of the instance of interest for those of the NUN. LASTS \cite{spinnato2023understanding} takes advantage of the growing sphere algorithm to create a latent space around the target instance, identifying the NUN. It builds a neighborhood around this NUN to generate a set of CEs. AB-CE \cite{li2023attention} uses a sliding window technique to generate multiple candidate subsequences, which are then evaluated based on Shannon Entropy to select the most informative ones. It then proceeds to swap the corresponding instance subsequences to be explained for the candidate subsequences one by one until the classification label changes. TSEvo \cite{hollig2022tsevo} formulates the search for CEs as a multi-objective optimization problem that incorporates proximity, sparsity, and plausibility. It initializes a population of candidate counterfactuals for a MTS instance and uses a customized version of NSGA-II \cite{deb2002fast} to evolve them. Individuals are evaluated based on the defined objectives, ranked using non-dominated sorting, and selected via tournament selection that considers both rank and crowding distance. Offspring are generated via crossover and mutation, repeated until the target number of generations is reached.

\definecolor{darkgreen}{RGB}{0, 100, 0}
\definecolor{mediumgreen}{RGB}{60,179,113}

\begin{table}[t]
\caption{
Summary of CE XAI methods for MTS. 
\textbf{Generation} columns show which objectives are optimized during CE generation, while \textbf{Evaluation} columns indicate which aspects are later assessed in the studies. 
Approach: He = Heuristic, Sh = Shapelet, At = Attention Mechanism, La = Latent space, Mo = Multi-objective. 
Scope: In = Instance-based, Su = Subsequence-based, Tp = Time-point based.
}
\centering\scalebox{.85}{
\begin{tabular}{p{1.5cm} p{0.4cm} p{0.4cm} | *{5}{p{0.2cm}} | *{5}{p{0.2cm}}} \toprule
 & & & \multicolumn{5}{c}{\textbf{Generation}} & \multicolumn{5}{|c}{\textbf{Evaluation}}  \\ 
\textbf{Method}  & \rotatebox{90}{\textbf{Approach}}   & \rotatebox{90}{\textbf{Scope}} & \rotatebox{90}{\textbf{Validity}} & \rotatebox{90}{\textbf{Confidence}} & \rotatebox{90}{\textbf{Sparsity}} & \rotatebox{90}{\textbf{Plausibility}} & \rotatebox{90}{\textbf{Proximity}} & \rotatebox{90}{\textbf{Validity}} & \rotatebox{90}{\textbf{Confidence}} & \rotatebox{90}{\textbf{Sparsity}} & \rotatebox{90}{\textbf{Plausibility}} & \rotatebox{90}{\textbf{Proximity}}\\ \midrule
CoMTE & He & In & &\cellcolor{darkgreen} & \cellcolor{darkgreen} & \cellcolor{darkgreen} & &  & \cellcolor{mediumgreen} & \cellcolor{mediumgreen} &  & \\
SETS & Sh & Su & &\cellcolor{darkgreen} &  & \cellcolor{darkgreen} &  &  & & \cellcolor{mediumgreen} & \cellcolor{mediumgreen} & \cellcolor{mediumgreen} \\
AB-CE & At & Su & &\cellcolor{darkgreen} &  & \cellcolor{darkgreen} &  & & \cellcolor{mediumgreen} & \cellcolor{mediumgreen} &  & \cellcolor{mediumgreen}\\
LASTS & La & Su & \cellcolor{darkgreen} & &  & \cellcolor{darkgreen} & \cellcolor{darkgreen} &  &  & & \cellcolor{mediumgreen} & \cellcolor{mediumgreen}\\
TSevo & Mo & Tp &  & & \cellcolor{darkgreen} & \cellcolor{darkgreen} & \cellcolor{darkgreen} & \cellcolor{mediumgreen} &  & \cellcolor{mediumgreen} & \cellcolor{mediumgreen} & \cellcolor{mediumgreen}\\
\textbf{\textsc{confetti}} & Mo & Tp & \cellcolor{darkgreen} & \cellcolor{darkgreen} & \cellcolor{darkgreen} & \cellcolor{darkgreen} & \cellcolor{darkgreen} & \cellcolor{mediumgreen}  & \cellcolor{mediumgreen} & \cellcolor{mediumgreen} & \cellcolor{mediumgreen} & \cellcolor{mediumgreen}\\
\bottomrule
\end{tabular}
}
\label{tab:summary_literature}
\end{table}

\paragraph{Limitations}
While all methods ensure Plausibility by design, only a few approaches (notably ours and LASTS) also enforce Validity by design. Beyond these requirements, most methods focus on a single optimization goal, either minimizing the distance between the original instance and its CE (e.g., LASTS, SETS) or maximizing the prediction confidence of the CE (e.g., CoMTE, AB-CE). Methods that attempt to address multiple criteria simultaneously, such as TSEvo and CoMTE, differ in how they explore the search space: TSEvo uses uninformed population-based exploration, making it costly and inefficient for high-dimensional or long time series, whereas CoMTE employs a more structured search, but still does not incorporate prior knowledge about the input space. To highlight the limitations of current methods, 

\figurename~\ref{fig:example} illustrates an example of CE generated by our method, CoMTE and TsEVO. The panel below \textsc{Confetti} shows the CAM of the NUN used to guide its changes. CoMTE and TSEvo largely replace the series with the NUN, resulting in low sparsity. By focusing on CAM-highlighted regions, \textsc{Confetti} achieves sparse and interpretable CEs. To address the limitations raised, we propose \textsc{Confetti}, which incorporates informed guidance through relevant subsequences and NUNs to perform the search process more effectively, drawing inspiration from CE techniques in other domains such as computer vision \cite{dandl2020multi,navas2021optimal}.

\section{CE Problem Definition}\label{sec:problem}
This section defines the core concepts and notation used in \textsc{confetti}.
%Intuitively, given a classifier and a set of MTS instances (some of which being classified as one class and others as other classes), our goal is to find a CE for a local instance, minimizing the differences between the original instance and its corresponding CE. 

\textbf{Basic Notation.}
Scalars are denoted by lowercase letters (e.g., $a, b, t, d, \alpha, \beta$). Sets are denoted by uppercase letters (e.g., $A$), and their elements by uppercase letters with subscripts (e.g., $A_i$). When the elements of a set are vectors (e.g., a timeseries),  we use superscripts to denote its elements (e.g., $A_i^t$ is the $\text{t}^{\text{th}}$ element of $A_i$). We use two comma-separated superscript indices to indicate a contiguous sequence of elements of a vector (e.g. $A_i^{s,e} = \{A_i^s,A_i^{s+1},\dots,A_i^e\}$). When each $A_i^t$ is itself a vector (e.g, a MTS), square brackets are used to select a specific element of it, so that $A_i^t[d]$ denotes the $d$‑th element of $A_i^t$.
We also define the operator $\mathbin{\|}$ to denote the concatenation of sequences. For example, $A_i \mathbin{\|} A_j = {A_i^0,A_i^1,  ..., A_i^{|A_i|}, A_j^0, A_j^1, ..., A_j^{|A_j|}}$. 

\begin{definition}[\textbf{Classifier, Instance, Prediction}]
A \textbf{classifier} $f$ is an estimator trained with MTS samples. We denote by $X_i \in X$ the $i^{\text{th}}$ \textbf{instance} of a set of MTSs $X$. A \textbf{prediction} $f(X_i)$ is the class predicted by the classifier~$f$ for $X_i$.
\end{definition}
Since an instance $X_i$ is a MTS, let us note that it can be sliced along its temporal and channel dimensions. 

\begin{definition}[\textbf{Subsequence}] 
A \textbf{subsequence}~$X^{s,e}_i \subseteq X_i$ is a contiguous segment of the original MTS instance $X_i$ from time $s$ to time $e$, so  $X^{s,e}_i = \{X_i^{s}, X_i^{s+1}, \dots, X_i^{e}\}$, with $0 \leq s < e < |X_i|$.  Each $X_i^{t}$ is a vector that contains the values of all channels at time $t$.
\end{definition}

\begin{definition}[\textbf{Nearest Unlike Neighbor - NUN}]
    Given a dataset~$X$, a classifier~$f$, a distance metric between two instances $d(X_a, X_b)$, and an instance $X_i \in X$, we denote the NUN of $X_i$ an instance $nun \in X~|~f(nun) \neq f(X_i)$, and $\nexists X_s \in X~|~f(X_s) \neq f(X_i) \text{ and } d(X_s,X_i) < d(nun, X_i)$.
    %Intuitively, the NUN is the nearest input that the model classifies differently from $X_i$; the distance $d(nun, X_i)$ therefore marks the smallest change in input space required to alter the model’s prediction.
\end{definition}
\begin{definition}[\textbf{Counterfactual Set, Counterfactual}]
    Given a dataset $X$, a classifier $f$ that outputs the prediction $f(X_i)$ for an instance $X_i \in X$, we denote by $C(X_i)$ the set of \textbf{counterfactuals} for $X_i$, where each $C_j \in C(X_i)$ satisfies $f(C_j) \neq f(X_i)$.
\end{definition}

%The intersection between an instance and its CE could be empty. However, it would mean that only a completely different instance could lead to a different prediction, which makes the usage of counterfactuals to explain the classifier's behavior ineffective. In our research, we assume the existence of counterfactuals that partially preserve the instance of interest. We also enforce that a CE $C_j$ is realistic or plausible, meaning it is not an adversarial example \cite{goodfellow2014explaining}, and that its values are consistent with the patterns and distributions present in the original dataset. 
To ensure that the CE values are within the data manifold, \textsc{confetti} -- \emph{as detailed in section~\ref{subsec:nun}} -- and many state-of-the-art methods construct counterfactuals by altering an instance~$X_i$ with values from its $nun$, thus promoting plausibility \cite{theissler2022explainable}.

\begin{definition}[\textbf{Prediction Confidence}]
Given a classifier~$f$, an instance~$X_{i}$, and a target class~$c$, we define the \textbf{prediction confidence} for class~$c$ as the probability~$P(f(X_{i}) = c)$ assigned by the classifier~$f$. This value quantifies the probability that the model considers class~$c$ to be the correct prediction for instance~$X_i$.
\end{definition}

\begin{definition}[\textbf{Minimality}]
Given an instance~$X_i$ and a CE~$C_j \in C(X_i)$, we define \textbf{minimality} as the distance~$d(X_i, C_j)$ between the original instance and its CE, according to a predefined distance metric~$d(\cdot,\cdot)$. A lower value of~$d$ indicates that the CE requires fewer or smaller changes to alter the prediction of the model, and therefore it is considered to be more minimal.
\end{definition}

%\begin{definition}[\textbf{Sparsity}]
%    Quantifies the number of components that remain unchanged between an original instance~$X$ and a CE~$C_j$. In the context of MTS, it measures how many time step values are identical across all channels. Higher sparsity indicates fewer changes, reflecting a more minimal and interpretable transformation. It is denoted by $\frac{1}{n} \sum_{i=1}^{n} \mathbb{I}(X_i = C_j)$
%\end{definition}

\paragraph{Problem Statement.}
Given a \textbf{classifier}~$f$, an \textbf{instance}~$X_i \in X$ with \textbf{predicted label}~$f(X_i)$, and a predefined distance metric~$d(\cdot, \cdot)$, the objective is to find a \textbf{set of counterfactuals} ~$C(X_i)$ such that for each \textbf{counterfactual instance} ~$C_j \in C(X_i)$ it holds that $f(C_j) \neq f(X_i)$, the distance~$d(X_i, C_j)$ is as \textbf{minimal} as possible, and the \textbf{prediction confidence}~$P(f(C_j) = c)$, for some class~$c \neq f(X_i)$, is as large as possible.

\section{CONFETTI}\label{sec:CONFETTI}
The process behind \textsc{confetti} consists of four main stages. First, a time series instance~$X_i$ is selected, and its $nun$ is identified as the counterfactual target. Second, the most influential subsequences affecting the prediction are identified. Third, values from $nun$ are substituted into $X_i$ to generate an initial CE that serves as the first element of the counterfactual set $C(X_i)$, denoted by $C_0$. Fourth, $C_0$ is passed to the optimization stage, which optimizes prediction confidence, proximity, and sparsity, producing an optimized set of CEs that belong to $C(X_i)$. These stages are respectively detailed in \sectionname s~\ref{subsec:nun} to \ref{subsec:optim}. We also outline the entire process in Algorithm \ref{alg:confetti}.

\subsection{Feature Weight Vector}
An optional component of \textsc{Confetti} is the use of a feature-weight vector $W_i$ to identify the most relevant contiguous subsequence within a MTS. This vector assigns an importance score to each time step, and subsequences with the highest cumulative weight are selected as candidates for modification. The weights can be obtained with techniques such as CAM \cite{zhou2016learning}, which requires a DL model with global average pooling to highlight key time steps by averaging feature maps before classification, enabling targeted rather than full-series perturbations. When a feature-weight vector $W_i$ is not available, for example due to the model architecture, \textsc{Confetti} skips the Subsequence extraction and Naive Stage steps (see \sectionname~\ref{subsec:subsequence-extraction} \& \ref{subsec:naive}) and proceeds directly to the optimization stage, in which case the method operates in a fully model-agnostic manner.

% Given a classifier~$f$ and a time series instance $X_i$, \textsc{confetti} returns a CE $C_j$ for $f(X_i)=f(X_i)$. Users can compare an instance with its set of CEs to identify which subsequences contribute most to the model's decision. They can either select a CE that balances the sparsity and prediction confidence, or choose to prioritize one objective/dimension.

% \textsc{confetti} works for DL models with a GAP layer due to the prevalence and advantages of such architectures in MTS classification. Most of the state-of-the-art DL models, like InceptionTime and its variants, are designed with a GAP layer as it significantly enhances model interpretability by linking feature maps to class scores. This layer also reduces the number of parameters, thus improving performance and mitigating overfitting. Furthermore, it has been demonstrated that CAMs, which rely on GAP layers, are highly effective in identifying the most relevant subsequences in time series data \cite{delaney2021instancebased}, allowing \textsc{confetti} to generate meaningful local CEs. 

\subsection{Retrieve the NUN}\label{subsec:nun}
We define $findNUN(\cdot)$ as the function that identifies the nearest unlike neighbor ($nun$) of a query instance $X_i$. It first filters the reference set $R$ to include only instances with a predicted label different from $f(X_i)$. A $k$-nearest neighbors search is then performed among these using a chosen distance metric. Candidates with classifier confidence above threshold $\theta$ are retained. If none meet this criterion, the function returns no neighbor and the process halts. Otherwise, it returns the closest valid $nun$ and its label $f(nun) = c$.

\begin{algorithm}[H]
\caption{Algorithm of \textsc{Confetti}}
\label{alg:confetti}
\textbf{Input:} Model $f$; instance $X_i$; reference data $R$; CAM weights of $R$ $W$; 
confidence weight $\alpha$; threshold $\theta$; partitions $p$; generations $g$; population size $z$; 
crossover probability $p_c$, mutation probability $p_m$

\begin{algorithmic}[1]
\State $(nun, c) \gets findNUN(f, X_i, R, \theta)$
\If{$nun = \bot$}
    \State \Return \textsc{Fail}
\Else
    \If{$\exists W$}
        \State $\ell \gets 2$
        \State $s,e \gets findSubsequence(W_{nun}, \ell)$
        \State $C_0 \gets X_i^{0,s-1} \mathbin{\|} nun^{s,e} \mathbin{\|} X_i^{e+1,|X_i|-1}$
        \While{$P(f(C_0) = c) < \theta$}
            \State $\ell \gets \ell + 1$
            \State $s,e \gets findSubsequence(W_{nun}, \ell)$
            \State $C_0 \gets X_i^{0,s-1} \mathbin{\|}  nun^{s,e} \mathbin{\|} X_i^{e+1,|X_i|-1}$
        \EndWhile
        \State $C(X_i) \gets \{C_0\}$
        \State $S(C(X_i)) \gets \{[0,0]\}$
        \State $h \gets \ell$
        
    \Else
        \State $C(X_i) \gets \emptyset$
        \State $S(C(X_i)) \gets \emptyset$
        \State $h \gets |X_i| / 2$
    \EndIf
\State $l \gets 1$

\While{$l \leq h$}
    \State $k \gets \left\lfloor \dfrac{l + h}{2} \right\rfloor$
    \State $s,e \gets findSubsequence(W_{nun}, k)$
    \State $H \gets referencePoints(2, p)$
    \State $P_0 \gets binarySampling(X_i, nun, s,e, z, H)$
    \State $evaluateObjectives(P_0, \alpha, \theta)$
    \State $P_g, S_g \gets nsgaIII(P_0,\ g,\ z,\ p_c,\ p_m,\,\ \theta)$
    \If{$P_g = \emptyset$}
        \State $l \gets k + 1$
    \Else
        \State $C(X_i) \gets C(X_i) \cup P_g$
        \State $S(C(X_i)) \gets S(C(X_i)) \cup S_g$
        \State $h \gets k - 1$
    \EndIf
\EndWhile

\State \Return $C(X_i)$, $\argmax{S(C(X_i)) \times [\alpha, 1-\alpha]}$
\EndIf
\end{algorithmic}
\end{algorithm}

\subsection{Subsequence extraction}\label{subsec:subsequence-extraction}
This step uses the classifier's feature-importance vector of $nun$, denoted by $W_{nun} = \{ W_{nun}^0, W_{nun}^1, \dots, W_{nun}^{|nun|-1} \}$, where $W_{nun}^t$ is the importance weight associated with time step $t$. This vector, obtained using CAM, is used to locate the most relevant subsequence of length $\ell$. A sliding window is applied to $W_{nun}$ to identify the contiguous segment with the largest cumulative weight, thereby restricting the perturbation to a smaller set of time steps. Although inspired by \cite{delaney2021instancebased}, \textsc{confetti} adapts this mechanism for MTS by averaging CAM values across channels, as implemented in the \textsc{findSubsequence}$(W_{nun}, \ell)$ function (Algorithm~\ref{alg:findsubsequence}).

\begin{algorithm}[t]
\caption{\textsc{findSubsequence}$(W_{nun}, \ell)$}
\label{alg:findsubsequence}
\textbf{Input:} Importance weights $W_{nun}$; subsequence length $\ell$ \\
\textbf{Output:} Indices $s, e$ such that $nun^{s,e}$ is the most relevant subsequence of length $\ell$

\begin{algorithmic}[1]
\State $s \gets 0$
\State $maxSum \gets \sum_{i=0}^{\ell - 1} W_{nun}^i$
\State $currSum \gets maxSum$
\For{$i = 1$ \textbf{to} $|W_{nun}| - \ell$}
    \State $currSum \gets currSum - W_{nun}^{[i - 1]} + W_{nun}^{[i + \ell - 1]}$
    \If{$currSum > maxSum$}
        \State $maxSum \gets currSum$
        \State $s \gets i$
    \EndIf
\EndFor
\State $e \gets s + \ell - 1$
\State \Return $s, e$
\end{algorithmic}
\end{algorithm}

%The identification of the most important subsequence for label prediction is central to many XAI methods for time series~\cite{ates2021counterfactual, spinnato2023understanding, li2023attention}. A critical subsequence can serve as an explanation itself, while also helping to minimize the modifications needed to generate CEs.

\subsection{Naive Stage}\label{subsec:naive}
In this step, the values of instance $X_i$ are altered by substituting them with the corresponding values from the same time steps in its $nun$. This stage is referred to as \textit{naive} because it applies the modification uniformly across all channels. Specifically, for a subsequence $X^{s,e}_i \subseteq X_i$ spanning a set of time steps, all values are replaced with those from the equivalent subsequence in $nun$, yielding the counterfactual instance $C_0$.

The process begins by setting the subsequence length $\ell = 2$, identifying the target class $f(nun) = c$, and selecting the most influential subsequence $nun^{s,e} \in nun$ based on CAM weights, with $e - s = \ell$. The corresponding segment $X^{s,e}_i$ in $X_i$ is then replaced with $nun^{s,e}$, creating a candidate $C_0$. The classifier’s confidence in predicting class $c$ for $C_0$, denoted $P(f(C_0) = c)$, is computed. If this value meets or exceeds the user-defined threshold $\theta$, the counterfactual is accepted. Otherwise, $\ell$ is incremented and the process repeats until the threshold is satisfied.

In the next stage, we input the optimization algorithm's final time window $(s,e)$; the optimization will occur within this window. In the most extreme scenario, the \textit{naive stage} algorithm will replace all values of $X_i$ with those of $nun$.

\subsection{Optimization}\label{subsec:optim}
\textsc{Confetti} aims to optimize the prediction confidence of the target class $c$, the distance between CE $C_j \in C(Xi)$ and $X_i$, and the number of changes made to a CE $C_j$. To evaluate the number of changes between CE $C_j \in C(Xi)$ and $X_i$, we use the Hamming distance \cite{hamming1950distance}:

\[
\mathrm{Hamming}(X_i, C_j) = \sum_{m=1}^{t} \sum_{n=1}^{u} \mathbf{1}_{\big[ X_{i}^m[n] \neq C_j^m[n] \big]}
\]

That is, the total number of elements changed between $X_i$ and $C_j$ in all $t$ time steps and $u$ channels

Our multi-objective optimization problem is then formalized as follows:
\begin{flalign}
\label{eq:optimization}
    \begin{split}
    \max \quad & m_1 =\sum_{C_i \in C(X_i)}{  P(f(C_j) = c)} \\
    \min \quad & m_2 = \frac{1}{t \cdot u} \cdot \sum_{C_i \in C(X_i)}^{}\mathrm{Hamming}(X_i, C_j) \\
    \min \quad & m_3 = \sum_{C_i \in C(X_i)}^{}\mathrm{dist}(X_i,C_j)\\
    \text{s.t.} \quad & P(f(C_j) = c) \geq \theta \text{,}\ \forall C_j \in C(X_i)\\
    \end{split}
\end{flalign}

In~\eqref{eq:optimization}, we maximize \( m_1 \), the model’s confidence, defined as the sum of predicted probabilities for class~\( c \) over all \( C_j \in C(X_i) \) under \( \mathit{f} \). The second objective, \( m_2 \), promotes sparsity by summing the distances between \( X_i \) and each \( C_j \), normalized by \( t \cdot u \), reflecting the proportion of changed elements across time and channels.  With \( m_3 \), we aim to minimize the distance between $X_i$ and each $C_j$. This distance can take the form of L1 norm, L2 norm or other distance metrics.  A constraint ensures that each \( C_j \in C(X_i) \) is classified as \( c \) by \( f \) with probability at least \( \theta \in [0,1] \).

If a solution is found within the time window, the subsequence length \( \ell \) is reduced by one, and the process repeats until no further solution is found. \textsc{Confetti} uses NSGA-III for optimization, chosen for its ability to maintain solution diversity via reference points generated using the Das-Dennis method~\cite{das1998normal} in a 3-objective space with \( k \) partitions. The initial population is generated via Binary Random Sampling, and the algorithm applies Two-Point Crossover and Bit-Flip Mutation. It runs for a fixed number of generations to evolve toward an optimal solution. Each generation produces a population $P_g$ and the values of the objectives of $S_g = \{ \forall x\in P_g | [m_1, m_2]\}$ that are added to the sets $C(X_i)$ and $S(C(X_i))$ respectively.

Lastly, \textsc{Confetti} includes a weighting parameter $\alpha \in [0,1] $ that allows users to balance the relative importance of confidence and sparsity when selecting the best CE. We combine these two objectives using a weighted sum, applying the vector $[\alpha, 1-\alpha]$. The CE $C_{j} \in C(X_{i})$ that obtains the highest weighted score is selected as the best candidate for instance $X_{i}$. Ultimately, \textsc{Confetti} returns the full set $C(X_i)$ and the best candidate.

\paragraph{Correctness of the approach.} Our method ensures that each \( C_j \in C(X_i) \) is generated through controlled optimization, progressively reducing modifications from the initial CE \( C_0 \in C(X_i) \). The following theorem shows that every \( C_j \) after the naive stage maintains or improves similarity to \( X_i \), measured by Hamming distance.

%Thus, the theorem provides a proof of our approach’s effectiveness in minimizing perturbations while achieving the desired confidence threshold.
\begin{theorem}
Let \( f \) be the classifier, \( X_i \) the instance to be explained, and \( \theta \in [0,1] \) a confidence threshold. Suppose \textsc{Confetti} is executed with these inputs and returns a set of counterfactuals \( C(X_i) \). Let \( C_0 \in C(X_i) \) denote the initial counterfactual generated during the naive stage (Lines 5–12). Define \( \mathrm{Hamming}(C_j, X_i) \) as the number of non-matching values between \( C_j \) and \( X_i \) across all time steps and channels—that is.

Then, for all $C_j \in C(X_i)$, it holds that $\mathrm{Hamming}(C_j, X_i) \leq \mathrm{Hamming}(C_0, X_i)$
\end{theorem}

\noindent\textit{Context.} 
The initial CE $C_0$ is constructed as 
$C_0 \gets X_i^{0,s-1} \mathbin{\|} nun^{s,e} \mathbin{\|} X_i^{e+1,|X_i|-1}$. 
If the replaced subsequence spans $\ell = e-s$ timesteps and the series has $d$ channels, 
then all $\ell \cdot d$ entries in that region differ from $X_i$, giving $
\mathrm{Hamming}(C_0, X_i) = \ell \cdot d$

\begin{proof} Throughout \textsc{Confetti}'s Algorithm (Algorithm \ref{alg:confetti}) the following properties hold:
\begin{enumerate}
  \item \textbf{Window bound.}\;%
        At every entry to the NSGA-III optimisation loop the current
        subsequence length satisfies $k \;\le\; \ell$
        
        \emph{Reason:} the outer loop sets
        $k \gets \lfloor \ell/2 \rfloor$; afterwards the binary-search
        update $h \gets k - 1$ can only \emph{decrease}
        the upper bound $\ell$.

  \item \textbf{Window locality.}\;%
        For a candidate $C_j$ produced by
        \textsc{binarySampling}, \textsc{crossover},
        or \textsc{mutation} we have
        \[
           \mathrm{Hamming}(C_j, X_i)\;\le\;k \cdot d .
        \]
        \emph{Reason:} These operators modify \emph{only} the current window $\text{X}_i^{s,e}$ of length $k$, affecting at most $k$ time steps and $k \cdot d$ time–channel entries.
\end{enumerate}

Now choose any CE $C_j$ that the algorithm finally returns.
Because $C_j$ comes from some generation of the NSGA-III loop,
Property~(2) yields $\mathrm{Hamming}(C_j, X_i) \leq k \cdot d$.
By Property~(1) we further have that $ k \leq \ell$; chaining the two inequalities, we obtain:
\[
   \mathrm{Hamming}(C_j, X_i) \leq \ell \cdot d
   = \mathrm{Hamming}(C_0, X_i).
\]

\end{proof}

\section{Experimental Evaluation}\label{sec:Exp}
To assess the ability of \textsc{confetti} to solve its problem statement, we focus on answering the following research questions in the experiments:
\begin{itemize}
    \item (RQ1)~Can \textsc{confetti} \emph{effectively} generate significant counterfactuals for MTS (\sectionname~\ref{sec:bench})?
    \item (RQ2)~Does \textsc{confetti}'s naive stage contributes significantly to the quality of the generated counterfactuals (\sectionname~\ref{sec:ablation})?
    \item (RQ3)~How \emph{sensitive} is \textsc{confetti} to the configuration of its internal parameters and components (\sectionname~\ref{sec:sensitivity})?
    \item (RQ4)~How \emph{efficient} is \textsc{confetti} compared to current alternatives in the state-of-the-art (\sectionname~\ref{sec:bench})?
\end{itemize}

%in multiple datasets and two distinct models, using a set of quantitative metrics introduced in this section. We compare against three state-of-the-art CE generation methods and examine how variations in parameters $\alpha$ and $\theta$ affect CE quality.

\paragraph{Models \& Datasets} 
The experiments are conducted with (i) two models: Fully Convolutional Network (FCN), Residual Network (ResNet) implemented in \cite{wang2017time}, which have demonstrated strong performance on MTS classification and support CAM extraction (\sectionname~\ref{sec:CONFETTI}); (ii) seven datasets from the Multivariate TSML Archive~\cite{bagnall2018uea} (see \tablename~\ref{tab:datasets}), selected to cover diverse time series lengths, channel counts, and class numbers.

\begin{table}[h]
\centering
\caption{Dataset Properties}
\small
\begin{tabular}{@{}lrrr@{}}
\toprule
\textbf{Dataset} & \textbf{Length} &  \textbf{Dimensions} & \textbf{Classes} \\ \midrule
ArticularlyWordRecognition & 144 & 9 & 25 \\ 
BasicMotions & 100 & 6 & 4 \\ 
Epilepsy  & 207 & 3 & 4 \\ 
ERing & 65 & 4 & 6 \\ 
Libras & 45 & 2 & 15 \\
NATOPS & 51 & 24 & 6 \\ 
RacketSports & 30 & 6 & 4 \\ 
\bottomrule
\end{tabular}
\vspace{0.5ex}
\parbox{0.9\linewidth}{\small \textit{Note.} We refer to ArticularlyWordRecognition as AWR in the subsequent text.}
\label{tab:datasets}
\end{table}

\paragraph{Evaluation criteria}
CE quality is assessed using six criteria and eight metrics, as follows:
\begin{enumerate}
    \item \textit{Sparsity \textbf{(SPA)}}: For each instance, measures the proportion of unchanged time–channel entries between a CE and its original, averaged over all instances:
        \[
            \frac{1}{|X_i|} \sum_{X_i \in X}
            \sum_{C_j \in C(X_i)}
            \left( 1-\frac{\mathrm{Hamming}(X_i, C_j)}{t \cdot d} \right)
        \]

   Dividing by the number of time steps \( t \) and channels \( d \) normalizes the Hamming distance to \([0,1]\); higher values indicate sparser counterfactuals.

    \item \textit{Counterfactual Confidence \textbf{(CONF)}}: For each instance, measures how much its counterfactuals reduce the model’s confidence in the original class \( f(X_i) \), averaged over all instances. Higher values indicate lower confidence in the original class:
\[
    \frac{1}{|X_i|} \sum_{X_i \in X}
    \sum_{C_j \in C(X_i)}
    \left( 1 - P(f(C_j) = f(X_i)) \right)
\]

\item \textit{Plausibility}: We use the \textbf{yNN} score~\cite{pawelczyk2021carla}, adapted to MTS as in~\cite{hollig2022tsevo}, to measure how well the predicted class of a CE~\( C_j \in C(X_i) \) is supported by its local neighborhood. For each \( C_j \), we compute its \( k = 5 \) nearest neighbors using DTW and average the yNN score across all instances. Higher values indicate stronger support from the data distribution.
\[
    \frac{1}{|X|} \sum_{X_i \in X} \sum_{C_j \in C(X_i)}
     1 - \frac{\sum_{R_h\in kNN(C_j)}
    \mathbf{1}_{\big[f(C_j)=f(R_{h})\big]}}{k}
\]

    \item \textit{Proximity}: Evaluated in three ways: (i) \( l_1 \) measures overall deviation between \( C_j \in C(X_i) \) and \( X_i \) without regard to location or size of changes; (ii) \( l_2 \) penalizes larger differences more heavily; (iii) DTW assesses temporal similarity. Lower values indicate better proximity.

\item \textit{Coverage \textbf{(COV)}}: Proportion of test instances \( X \) for which at least one CE is found (the higher the better):
\[
    \frac{1}{|X|} \sum_{X_i \in X} \mathbf{1}_{C(X_i) \neq \emptyset}
\]

\item \textit{Validity \textbf{(VAL)}}: Proportion of CEs that successfully change the original prediction:
\[
    \frac{1}{|X|} \sum_{X_i \in X}
    \frac{1}{|C(X_i)|} \sum_{C_j \in C(X_i)}
    \mathbf{1}_{\big[ f(X_i) \neq f(C_j) \big]}
\]
\end{enumerate}
\paragraph{Baselines}
\textsc{Confetti} is compared with three state-of-the-art methods described in \sectionname~\ref{sec:LR}: CoMTE~\cite{ates2021counterfactual}, SETS~\cite{bahri2022shapelet}, and TSEvo~\cite{hollig2022tsevo}. We did not include AB‑CE \cite{li2023attention} due to the lack of publicly available code, which prevented a reproducible implementation. LASTS \cite{spinnato2022explaining} was also excluded because it produces multiple types of explanations (counterfactuals, rules, and a surrogate model) rather than focusing solely on counterfactuals, making a fair and direct comparison with our counterfactual-only evaluation framework impractical. 

The parameter configurations are as follows: CoMTE is configured to use one distractor, with a maximum of $100$~attempts and $100$~iterations; SETS is configured with a minimum shapelet length defined as either three time steps or one-tenth of the series length (whichever is greater), and a maximum shapelet length set to either half of the series length or one time step longer than the minimum (whichever is greater), ensuring shapelets scale appropriately with series size; finally, TSEvo is limited to a maximum of 100 epochs. All remaining parameters were kept at their default values provided by \texttt{TSInterpret}.

\paragraph{Implementation Setup}
\textsc{Confetti} was implemented in Python~3.12 using \texttt{Keras}~3.8.0 for modeling and \texttt{sktime}~0.36.0 for data handling. Baselines were built with \texttt{TSInterpret}~0.4.7. Experiments ran on a MacBook Pro (Apple M4 Max, macOS Sequoia~15.5, 36GB RAM). We used the official UEA train/test splits from \texttt{sktime}. Minor edits were made to \texttt{TSInterpret} to fix a CoMTE threading bug.

\subsection{Benchmark}\label{sec:bench}
To ensure fair comparison, we first evaluate the two metrics optimized by \textsc{Confetti}, limiting comparisons to methods that explicitly target each metric (\tablename~\ref{tab:summary_literature}). We compare \textit{Counterfactual Confidence} with CoMTE and SETS, and \textit{Sparsity} with CoMTE and TSEvo. We then report all eight metrics across all methods. Finally, we assign ranks (in parentheses) via pairwise method comparisons per dataset. A method ranks higher only if it outperforms another on every dataset (excluding missing values), with statistical significance confirmed by a paired Wilcoxon signed-rank test~\cite{wilcoxon1992individual} at \(\alpha = 0.05\); otherwise, ranks are shared.

%%%% Counterfactual Confidence %%%%

\subsubsection{Counterfactual Confidence}\label{sec:confidence}
\tablename~\ref{tab:confidence_scores} presents the benchmark results for \textit{Counterfactual Confidence}. Across all seven datasets, \textsc{Confetti} consistently outperforms CoMTE and SETS. With a stricter threshold (\( \theta = 0.95 \)), it achieves a mean confidence of 0.98, exceeding the best baseline (CoMTE, 0.86) by 0.02–0.22 across datasets—most notably on NATOPS, improving from 0.76 to 0.98.

\begin{table}[ht]
\centering
\caption{Confidence scores across methods for each dataset. Values are averaged over models. Ranks appears in parentheses in the column headers}
\resizebox{\linewidth}{!}{
\begin{tabular}{l cc cc}
\toprule
\multicolumn{1}{c}{} & \multicolumn{2}{c}{} & \multicolumn{2}{c}{\textsc{Confetti}} \\
\textbf{Dataset} 
& \makecell[b]{CoMTE\\(2)}
& \makecell[b]{SETS\\(2)}
& \makecell[b]{$\alpha$=0.5\\$\theta$=0.51\\(2)}
& \makecell[b]{$\alpha$=0.5\\$\theta$=0.95\\(1)} \\
\midrule
ARW            & 0.953 & 0.940 & 0.726 & \textbf{0.978} \\
BasicMotions   & 0.917 & 0.487 & 0.611 & \textbf{0.965} \\
ERing          & 0.701 & 0.766 & 0.770 & \textbf{0.981} \\
Epilepsy       & 0.837 & 0.778 & 0.636 & \textbf{0.972} \\
Libras         & 0.952 & 0.820 & 0.719 & \textbf{0.973} \\
NATOPS         & 0.755 & *     & 0.709 & \textbf{0.976} \\
RacketSports   & 0.932 & 0.780 & 0.756 & \textbf{0.980} \\
\bottomrule
\end{tabular}
}
\label{tab:confidence_scores}
\makebox[\linewidth][l]{\parbox{\linewidth}{\scriptsize\textit{Rank 1 values are highlighted in bold.\\ * Failed to produce results.}}}
\end{table}

%%% Sparsity %%%%
\begin{table}[ht]
\centering
\caption{Sparsity scores across methods for each dataset. Values are averaged over models. Ranks appears in parentheses in the column headers.}
\resizebox{\linewidth}{!}{
\begin{tabular}{l cc cc}
\toprule
\multicolumn{1}{c}{} & \multicolumn{2}{c}{} & \multicolumn{2}{c}{\textsc{Confetti}} \\
\textbf{Dataset} 
& \makecell[b]{CoMTE\\(3)} 
& \makecell[b]{TSEvo\\(4)}
& \makecell[b]{$\alpha$=0.0\\$\theta$=0.51\\(1)}
& \makecell[b]{$\alpha$=0.5\\$\theta$=0.51\\(2)} \\
\midrule
ARW            & 0.731 & 0.002 & \textbf{0.926} & 0.912 \\
BasicMotions   & 0.486 & 0.003 & \textbf{0.822} & 0.798 \\
ERing          & 0.681 & 0.029 & \textbf{0.913} & 0.876 \\
Epilepsy       & 0.461 & 0.011 & \textbf{0.822} & 0.802 \\
Libras         & 0.247 & 0.033 & \textbf{0.850} & 0.794 \\
NATOPS         & 0.719 & 0.001 & \textbf{0.880} & 0.861 \\
RacketSports   & 0.562 & 0.012 & \textbf{0.942} & 0.912 \\
\bottomrule
\end{tabular}
}
\label{tab:sparsity_scores}
\makebox[\linewidth][l]{\scriptsize\textit{Note: Rank 1 values are highlighted in bold.}}
\end{table}

The default setting (\( \theta = 0.51 \)) further demonstrates \textsc{Confetti}’s adaptability. While its mean confidence (0.70) is slightly below CoMTE’s, it remains competitive on four datasets and clearly outperforms SETS on BasicMotions, ERing, Epilepsy, and RacketSports. Thus, \( \theta = 0.95 \) is preferable for high-stakes scenarios requiring high reliability, whereas \( \theta = 0.51 \) offers a solid baseline for general use.

%%%% Sparsity %%%%
\subsubsection{Sparsity}\label{sec:sparsity}
\tablename~\ref{tab:sparsity_scores} reports results for \textit{sparsity}, with ranks assigned as in \tablename~\ref{tab:confidence_scores}. \textsc{Confetti} again outperforms both CoMTE and TSEvo on all datasets. With full emphasis on sparsity (\( \alpha = 0.0 \)), it reaches an average score of 0.88, clearly ahead of CoMTE (0.56) and TSEvo (0.01). Even the balanced setting (\( \alpha = 0.5 \)) performs strongly, averaging 0.85. These results show that \textsc{Confetti} offers precise control over the sparsity–confidence trade-off, with \( \alpha = 0.0 \) suited for minimizing feature changes and \( \alpha = 0.5 \) as a robust general setting.

\subsubsection{Evaluation Metrics Overview}\label{sec:results_all}
\tablename~\ref{tab:fcn_scores} and~\ref{tab:resnet_scores} report average scores for all methods using FCN and ResNet, along with ranking (in parenthesis). Best-performing methods are shown in bold. Multiple bold entries indicate no significant difference.

All \textsc{Confetti} variants consistently achieved 100\% coverage (COV), matching CoMTE and TSEvo, meaning CEs were generated for every instance. In terms of validity (VAL), \textsc{Confetti} outperformed TSEvo and SETS and, while tied in rank with CoMTE, was the only method to reach a perfect score of 1 across both models and all datasets, indicating every CE changed the model prediction. By contrast, CoMTE ranged between 0.79–0.80 and 0.91–0.93.

\begin{table}[ht]
\caption{Performance of CE methods on FCN for all evaluation metrics. Values represent average scores across all datasets. Ranking position are in parentheses}
\centering
\scriptsize
\setlength{\tabcolsep}{4.1pt}
\begin{tabularx}{\columnwidth}{lccc rcl}
\toprule
\multicolumn{1}{c}{} & \multicolumn{3}{c}{Baselines} & \multicolumn{3}{c}{\textsc{Confetti}} \\
\textbf{Metric} & CoMTE & SETS & TsEVO & \phantom{ab}
\makecell{$\alpha$=0.0\\$\theta$=0.51} &
\makecell{$\alpha$=0.5\\$\theta$=0.51} &
\makecell{$\alpha$=0.5\\$\theta$=0.95} \\
\midrule
COV     & \textbf{100 (1)} & \textbf{94.17 (1)} & \textbf{100 (1)} & \textbf{100 (1)} & \textbf{100 (1)} & \textbf{100 (1)} \\
VAL     & \textbf{0.93 (1)} & 0.77 (2) & 0.80 (2) & \textbf{1.00 (1)} & \textbf{1.00 (1)} & \textbf{1.00 (1)} \\
CONF   & 0.86 (2) & 0.76 (3) & 0.80 (2) & 0.59 (4) & 0.69 (3) & \textbf{0.97 (1)} \\
SPA     & 0.54 (4) & 0.02 (5) & 0.01 (5) & \textbf{0.88 (1)} & 0.85 (2) & 0.81 (3) \\
$L_1$        & 283.55 (4) & 923.58 (5) & 954.54 (5) & \textbf{99.76 (1)} & 111.80 (2) & 146.19 (3) \\
$L_2$        & 27.24 (4) & 56.79 (5) & 56.50 (5) & \textbf{16.01 (1)} & 16.83 (2) & 19.78 (3) \\
DTW          & 26.49 (4) & 48.16 (5) & 49.19 (5) & \textbf{15.40 (1)} & 16.09 (2) & 18.93 (3) \\
yNN          & \textbf{0.99 (1)} & \textbf{0.99 (1)} & \textbf{0.99 (1)} & \textbf{0.99 (1)} & \textbf{0.99 (1)} & \textbf{0.99 (1)} \\
\bottomrule
\end{tabularx}
\label{tab:fcn_scores}
\end{table}

\begin{table}[ht]
\caption{Performance of CE methods on ResNet for all evaluation metrics. Values represent average scores across all datasets. Ranking position are in parentheses.}
\centering
\scriptsize
\setlength{\tabcolsep}{4.1pt}
\begin{tabularx}{\columnwidth}{lccc rcl}
\toprule
\multicolumn{1}{c}{} & \multicolumn{3}{c}{Baselines} & \multicolumn{3}{c}{\textsc{Confetti}} \\
\textbf{Metric} & CoMTE & SETS & TsEVO & \phantom{ab}
\makecell{$\alpha$=0.0\\$\theta$=0.51} &
\makecell{$\alpha$=0.5\\$\theta$=0.51} &
\makecell{$\alpha$=0.5\\$\theta$=0.95} \\
\midrule
COV     & \textbf{100 (1)} & \textbf{93.24 (1)} & \textbf{100 (1)} & \textbf{100 (1)} & \textbf{100 (1)} & \textbf{100 (1)} \\
VAL     & \textbf{0.91 (1)} & 0.76 (2) & 0.79 (2) & \textbf{1.00 (1)} & \textbf{1.00 (1)} & \textbf{1.00 (1)} \\
CONF   & \textbf{0.87 (1)} & 0.76 (2) & 0.79 (2) & 0.59 (3) & 0.71 (2) & \textbf{0.98 (1)} \\
SPA     & 0.57 (4) & 0.02 (5) & 0.01 (5) & \textbf{0.88 (1)} & 0.85 (2) & 0.82 (3) \\
$L_1$        & 269.82 (4) & 920.46 (5) & 954.39 (5) & \textbf{96.50 (1)} & 106.08 (2) & 121.78 (3) \\
$L_2$        & 25.90 (3) & 56.98 (4) & 56.22 (4) & \textbf{14.94 (1)} & 15.61 (2) & 16.47 (3) \\
DTW          & 25.25 (4) & 47.72 (5) & 48.62 (5) & \textbf{14.22 (1)} & 14.79 (2) & 15.68 (3) \\
yNN          & \textbf{0.99 (1)} & \textbf{0.99 (1)} & \textbf{0.99 (1)} & \textbf{0.99 (1)} & \textbf{0.99 (1)} & \textbf{0.99 (1)} \\
\bottomrule
\end{tabularx}
\label{tab:resnet_scores}
\end{table}

Regarding sparsity, \textsc{Confetti} clearly outperforms all baselines. All tested configurations achieved significantly higher sparsity scores (0.81–0.88) than CoMTE (~0.56), and vastly exceeded SETS and TSEvo (both $< 0.02$). Sparsity remained consistently high across configurations, confirming \textsc{Confetti}’s consistently in generating minimal yet effective CEs.

\textsc{Confetti} also excelled in proximity metrics (\( l_1 \), \( l_2 \), and \( \mathrm{DTW} \)), achieving the lowest average distances across all settings and outperforming baselines by wide margins. Notably, proximity was not explicitly optimized, yet the CEs remained close to the original \( X_i \), enhancing interpretability. Since values are either retained from \( X_i \) or replaced with those from \( nun \), similar numbers of changes across methods result in comparable \( \mathrm{DTW} \) distances.

Finally, all methods (incl., \textsc{Confetti}) achieve high yNN scores (0.99), confirming plausibility within the data distribution. Together, these findings answer \textbf{RQ1} and \textbf{RQ4} positively: \textsc{Confetti} consistently generates valid, interpretable counterfactuals for MTS (RQ1) and outperforms or matches all baselines across metrics and datasets, demonstrating strong practical value (RQ4).

\subsubsection{Execution Times}\label{subsub:exec_times}
Lastly, we also show the execution times of the different methods across both models. We present our results in \tablename~\ref{tab:execution_times}. 

\begin{table}[ht]
\caption{Execution time comparison (in seconds) of CE methods for FCN and ResNet. Values represent mean and standard deviation across all datasets.}
\centering
\scriptsize
\setlength{\tabcolsep}{5.5pt}
\begin{tabular}{lcc}
\toprule
\textbf{Methods} & \textbf{FCN} & \textbf{ResNet} \\
\midrule
CoMTE                & 341.24 $\pm$ 77.52  & 308.73 $\pm$ 93.42 \\
\textsc{Confetti} $\alpha=0.0$ & 204.47 $\pm$ 118.07 & 335.37 $\pm$ 201.95 \\
\textsc{Confetti} $\alpha=0.5$ & 199.55 $\pm$ 115.59 & 339.00 $\pm$ 208.56 \\
\textsc{Confetti} $\theta=0.95$ & 223.85 $\pm$ 120.57 & 368.93 $\pm$ 221.73 \\
SETS                 & \textbf{84.01 $\pm$ 88.48}    & \textbf{89.02 $\pm$ 95.13} \\
TsEVO                & 8625.38 $\pm$ 2509.38 & 9126.82 $\pm$ 2615.85 \\
\bottomrule
\end{tabular}
\label{tab:execution_times}
\end{table}

As shown in the table, SETS is approximately an order of magnitude faster than the other methods, while TsEVO is roughly an order of magnitude slower. \textsc{Confetti} and CoMTE fall in the middle with comparable execution times, although CoMTE exhibits much lower variance across datasets. In contrast, \textsc{Confetti} shows a larger standard deviation, which is expected given its optimization process. The window-based search depends on the length and structure of each time series, making execution time more sensitive to dataset characteristics. This reflects a broader trade-off between methods that operate at the subsequence or timestep level and those that work at the channel level. Nevertheless, the fact that our subsequence-based approach remains competitive with CoMTE and significantly outperforms its most similar competitor TsEVO highlights the strength and practical relevance of \textsc{Confetti}.

\subsection{Ablation Study}\label{sec:ablation}
To respond to RQ2 we performed an ablation study of the naive stage of \textsc{confetti}. We carried out this experiment using the default hyperparameter settings ($\alpha$=0.5, $\theta=0.51$) in two different versions: one in which we provided CAM weights and the other in which we did not provide them. We evaluated the generated CE across all metrics and datasets and performed a Wilcoxon test \cite{wilcoxon1992individual} with Holm correction to identify whether there are any significant differences for each metric. We present our results in \tablename~\ref{tab:ablation_scores}.

\begin{table}[ht]
\caption{Performance comparison of \textsc{confetti} with and without Naive Stage for FCN and ResNet. Results are averaged across all datasets. A parenthesis (*) indicates a statistically significant difference (Wilcoxon + Holm).}
\centering
\scriptsize
\setlength{\tabcolsep}{5.5pt}
\begin{tabular}{lcc|cc}
\toprule
 & \multicolumn{2}{c}{\textbf{FCN}} & \multicolumn{2}{c}{\textbf{ResNet}} \\
\textbf{Metric} & Normal & Ablation & Normal & Ablation \\
\midrule
COV     & 100.00    & 99.40     & 100.00    & 100.00    \\
SPA     & 0.84      & 0.81      & 0.84      & 0.80      \\
CONF    & 0.58      & 0.58      & 0.55      & 0.55      \\
VAL     & 1.00      & 1.00      & 1.00      & 1.00      \\
$L_{1}$ & 107.01 (*) & 145.22 (*) & 110.74 (*) & 168.13 (*) \\
$L_{2}$ & 15.65 (*)  & 19.61 (*)  & 15.69 (*)  & 20.59 (*)  \\
DTW     & 14.96 (*)  & 18.98 (*)  & 14.87 (*)  & 19.84 (*)  \\
yNN     & 0.99 & 0.99 & 0.99 & 0.99 \\
Exec. Time  & 162.76 (*) & 221.44 (*) & 302.84 (*) & 384.31 (*) \\
\bottomrule
\end{tabular}
\label{tab:ablation_scores}
\end{table}

First, we noticed that we did not find significant differences in coverage, sparsity, confidence, validity, and plausibility. This result is beneficial because we prove that \textsc{confetti} can achieve great results for models that do not have any feature importance mechanisms, and thus its application is relevant as a model-agnostic method. 
However, we found significant differences in performance across all proximity distance metrics, meaning that CEs produced with the aid of the naive stage resemble closer to the original instances, proving that using feature importance weights has indeed a positive impact. Furthermore, we also observed a significant difference in execution time. This improvement comes from the reduced size of the initial optimization windows. In the naive stage, the largest subsequence is located much more quickly, whereas the ablation variant always begins its optimization from a subsequence length fixed at half of the time series, resulting in a noticeably slower search.

\subsection{Sensitivity Analysis}\label{sec:sensitivity}
\begin{figure}[ht]
  \centering
  \includegraphics[width=\linewidth]{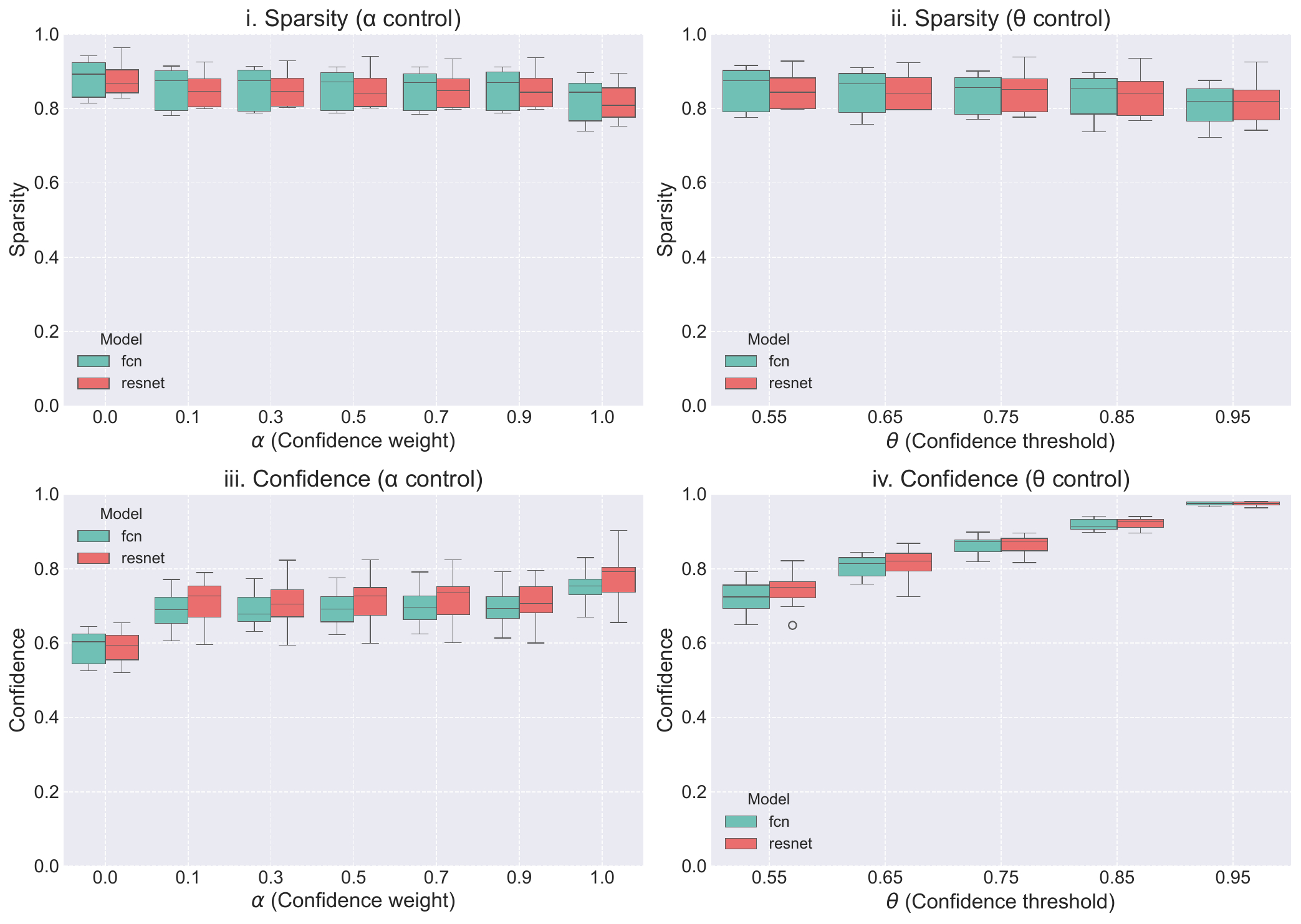}
 \caption{Effect of parameter configurations on counterfactual generation. Panels show: (i) Sparsity as $\alpha$  varies; (ii) Sparsity as $\theta$ varies; (iii) Confidence as $\alpha$ varies; and (iv) Confidence as $\theta$ varies.}
\label{fig:tradeoffs}
\end{figure}

To evaluate the impact of parameters $\theta$ and $\alpha$, we conducted two sensitivity analyses: one varying $\alpha$ with fixed $\theta$=0.51, and another varying $\theta$ with fixed $\alpha$=0.5. Specifically, we tested \( \alpha \in [0.0, 0.1, 0.3, 0.5, 0.7, 0.9, 1.0] \) and \( \theta \in [0.55, 0.65, 0.75, 0.85, 0.95] \). \figurename~\ref{fig:tradeoffs} shows how Sparsity, Confidence, and Proximity metrics respond to these parameter changes.

\subsubsection{Balancing Optimization Objectives}
We first examine the effect of \( \alpha \), which balances prediction confidence and sparsity during optimization. Lower \( \alpha \) favors sparsity; higher values prioritize confidence.

\textsc{Confetti}'s sparsity remains stable for intermediate \( \alpha \) values, with notable changes only at the extremes (\( \alpha = 0 \) or \( 1 \)). Thus, the optimization is sensitive to \( \alpha \) mainly when it is pushed to its limits.

% We observe a similar trend for \textit{proximity}, confirming our observation that it is a proxy for \textit{sparsity}. 

\subsubsection{Effect of Confidence Requirement}
\figurename~\ref{fig:tradeoffs} (subplot~iii) shows that increasing \( \alpha \) steadily improves \textit{confidence}, from just above 0.60 to $\approx 0.80$ at \( \alpha = 1 \). A similar trend appears in the sensitivity to \( \theta \): sparsity remains stable, while confidence increases with higher thresholds. Even at the lowest setting (\( \theta = 0.55 \)), \textsc{Confetti} produces CEs averaging $\approx 0.75$ in confidence, well above the required minimum. These results suggest that boosting prediction confidence generally requires modifying only a few time steps, reinforcing the benefit of CAM-guided perturbations. Notably, \( \alpha \) has no effect on \textit{Coverage}: \textsc{Confetti} generates CEs for all instances across datasets, regardless of the weighting. This invariance is by design, as the CE search process is independent of \( \alpha \). Similarly, \textsc{Confetti}'s consistency to increasing \( \theta \) likely reflects properties of the classifier \( f \), which yields a valid \( nun \) for every \( X_i \). These findings answer \textbf{RQ3}: \textsc{Confetti} is robust to parameter choices. \( \theta \) influences confidence but not sparsity, and \( \alpha \) only affects outcomes at extremes. Across a broad range of settings, \textsc{Confetti} reliably generates confident, sparse, and valid CEs with minimal tuning.

\subsection{Fitness Function Analysis}\label{subsec:fitnessanalysis}

\begin{figure}[ht]
  \centering
  \includegraphics[width=\linewidth]{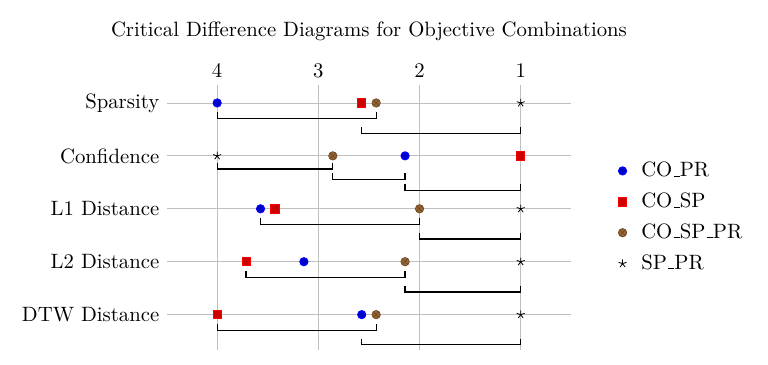}
 \caption{Critical Difference Diagram showing the average ranking (lower is better) of different objective combinations on counterfactual generation metrics. The lines group combinations that are not significantly different from each other. The different combinations are i) Confidence-Proximity (CO\_PR); ii) Confidence-Sparsity (CO\_SP); iii) Confidence-Sparsity-Proximity (CO\_SP\_PR); and iv) Sparsity-Proximity (SP\_PR).}
\label{fig:objectives}
\end{figure}

We conducted an empirical experiment to verify that our proposed fitness function was the right approach. To this end, we executed \textsc{confetti}'s CE search using all combinations of objectives. The tested configurations were Confidence-Proximity (CO\_PR), Confidence-Sparsity (CO\_SP), Sparsity-Proximity (SP\_PR), and Confidence-Sparsity-Proximity (CO\_SP\_PR). After evaluating all resulting CEs, we applied the Friedman test to assess whether ranking differences were statistically significant. Whenever the Friedman test \cite{friedman1937use} indicated significance at p < 0.05, we followed with the Nemenyi test \cite{nemenyi1963distribution} to determine which specific rankings differed. Only configurations with a Nemenyi p value below 0.05 were considered significantly different. The corresponding critical difference diagram is shown in \figurename~\ref{fig:objectives}.

As shown in \figurename~\ref{fig:objectives}, we first note that we do not report coverage, validity, or plausibility. We found no significant ranking differences for these metrics; all configurations reached identical scores (100\%, 1.0, and 0.99). For sparsity, we observed a significant difference between CO\_PR and SP\_PR, driven by their consistent ranking of 4th and 1st respectively across all datasets. This is expected since SP\_PR is the only configuration that does not incorporate confidence. However, once sparsity is included in the objective function (e.g., CO\_SP), we no longer observe a significant difference with respect to SP\_PR.

For confidence, we detected significant differences between CO\_SP and SP\_PR, as well as between the full configuration CO\_SP\_PR and SP\_PR, but not between CO\_PR and SP\_PR. This indicates that pairing confidence exclusively with sparsity or proximity does not substantially affect confidence performance, whereas including all three objectives leads to a slight decrease.

Regarding proximity metrics, we observe a consistent pattern in which SP\_PR differs significantly from both CO\_SP and CO\_PR (with the exception of DTW distance). However, no significant differences are found once all three objectives are combined in the fitness function.

At first glance, SP\_PR might appear to be the best option, but its poor confidence performance makes it unsuitable, given the importance of confidence in CE desiderata. The next best configurations for confidence were CO\_PR and CO\_SP\_PR. Between these two, CO\_PR is significantly different from SP\_PR in sparsity. Overall, our results indicate that including all three objectives provides the most balanced alternative. Although CO\_SP\_PR never achieved the top rank, it consistently remained close enough that its differences were not statistically significant.

\subsection{Choice of Proximity Distance}
As discussed in \sectionname~\ref{subsec:optim}, the objective $m_3$ can be instantiated with any distance function. To assess whether the choice of distance metric influences performance, we ran \textsc{confetti} using several alternatives: Euclidean distance, Canonical Time Warping (CTW) \cite{zhou2009canonical}, Dynamic Time Warping (DTW) \cite{sakoe2003dynamic}, and Soft-DTW \cite{cuturi2017soft}. Following the same procedure described in \sectionname~\ref{subsec:fitnessanalysis}, we applied a Friedman test to the resulting performance metrics to evaluate whether any metric led to statistically different rankings. In this analysis, we found no significant differences across any proximity metric for any of the CE performance measures. The results are presented in \tablename~\ref{tab:distance_scores}.

\begin{table}[ht]
\caption{Performance of \textsc{confetti} given different distance metrics for proximity for all evaluation metrics. Values represent mean and standard deviation across all datasets.}
\centering
\scriptsize
\setlength{\tabcolsep}{4.1pt}
\begin{tabularx}{\columnwidth}{lcccc}
\toprule
\textbf{Metric} & \textbf{ctw} & \textbf{dtw} & \textbf{euclidean} & \textbf{softdtw} \\
\midrule
COV   & 100.00 $\pm$ 0.00 & 100.00 $\pm$ 0.00 & 100.00 $\pm$ 0.00 & 100.00 $\pm$ 0.00 \\
VAL   & 1.00 $\pm$ 0.00 & 1.00 $\pm$ 0.00 & 1.00 $\pm$ 0.00 & 1.00 $\pm$ 0.00 \\
SPA   & 0.85 $\pm$ 0.05 & 0.85 $\pm$ 0.05 & 0.84 $\pm$ 0.07 & 0.84 $\pm$ 0.06 \\
CONF  & 0.59 $\pm$ 0.08 & 0.59 $\pm$ 0.08 & 0.58 $\pm$ 0.05 & 0.59 $\pm$ 0.09 \\
$L_1$ & 120.97 $\pm$ 133.22 & 120.97 $\pm$ 133.22 & 101.99 $\pm$ 110.69 & 120.55 $\pm$ 134.17 \\
$L_2$ & 17.41 $\pm$ 18.62 & 17.41 $\pm$ 18.62 & 15.08 $\pm$ 15.40 & 17.46 $\pm$ 18.69 \\
DTW   & 15.38 $\pm$ 17.37 & 15.38 $\pm$ 17.37 & 14.47 $\pm$ 15.20 & 15.59 $\pm$ 17.59 \\
yNN   & 0.99 $\pm$ 0.01 & 0.99 $\pm$ 0.01 & 0.99 $\pm$ 0.00 & 0.99 $\pm$ 0.01 \\
\bottomrule
\end{tabularx}
\label{tab:distance_scores}
\end{table}

\section{Conclusion}\label{sec:concl}
We introduced a novel method for generating CEs for MTS that effectively balances sparsity, proximity, and prediction confidence while ensuring plausibility and validity by design. Our approach leverages prior knowledge through CAM as an attention mechanism, enabling modifications of the most relevant subsequences. The experimental results confirmed that our method successfully generated valid CEs for all instances tested. Furthermore, it significantly outperformed existing state-of-the-art methods in sparsity, proximity, and also in counterfactual confidence, particularly at high values of the parameter $\theta$. Sensitivity analyzes demonstrate that modifying even a small number of time steps within the most relevant subsequence can substantially enhance prediction confidence while maintaining high sparsity. Lastly, our ablation study shows that \textsc{confetti} maintains almost the same performance even when attention mechanisms are removed, which confirms that the method can operate in a model-agnostic setting. However, this comes at the cost of worse proximity values and longer execution times.

\section{Acknowledgements}
This work is supported in whole or in part by the National Research Fund, Luxembourg. Grant references 18959931 and 18435508. The authors acknowledge the contributors to the UEA and UCR Time Series Repositories. Their work in creating, curating, and sharing these datasets has been essential for this research.

\bibliography{aaai2026}

@String{Computing = "Computing" }

@String{Computer = "{IEEE} Computer" }

@String{Springer = "Springer-Verlag" }

@article{ismail2019deep,
  title={Deep learning for time series classification: a review},
  author={Ismail Fawaz, Hassan and Forestier, Germain and Weber, Jonathan and Idoumghar, Lhassane and Muller, Pierre-Alain},
  journal={Data mining and knowledge discovery},
  volume={33},
  number={4},
  pages={917--963},
  year={2019},
  publisher={Springer}
}

@article{ruiz2021great,
  title={The great multivariate time series classification bake off: a review and experimental evaluation of recent algorithmic advances},
  author={Ruiz, Alejandro Pasos and Flynn, Michael and Large, James and Middlehurst, Matthew and Bagnall, Anthony},
  journal={Data Mining and Knowledge Discovery},
  volume={35},
  number={2},
  pages={401--449},
  year={2021},
  publisher={Springer}
}

@article{theissler2022explainable,
  title={Explainable AI for time series classification: a review, taxonomy and research directions},
  author={Theissler, Andreas and Spinnato, Francesco and Schlegel, Udo and Guidotti, Riccardo},
  journal={IEEE Access},
  volume={10},
  pages={100700--100724},
  year={2022},
  publisher={IEEE}
}

@article{adadi2018peeking,
  title={Peeking inside the black-box: a survey on explainable artificial intelligence (XAI)},
  author={Adadi, Amina and Berrada, Mohammed},
  journal={IEEE access},
  volume={6},
  pages={52138--52160},
  year={2018},
  publisher={IEEE}
}

@inproceedings{spinnato2022explaining,
  title={Explaining crash predictions on multivariate time series data},
  author={Spinnato, Francesco and Guidotti, Riccardo and Nanni, Mirco and Maccagnola, Daniele and Paciello, Giulia and Farina, Antonio Bencini},
  booktitle={International Conference on Discovery Science},
  pages={556--566},
  year={2022},
  organization={Springer}
}

@inproceedings{dandl2020multi,
  title={Multi-objective counterfactual explanations},
  author={Dandl, Susanne and Molnar, Christoph and Binder, Martin and Bischl, Bernd},
  booktitle={International Conference on Parallel Problem Solving from Nature},
  pages={448--469},
  year={2020},
  organization={Springer}
}

@inproceedings{ates2021counterfactual,
  title={Counterfactual explanations for multivariate time series},
  author={Ates, Emre and Aksar, Burak and Leung, Vitus J and Coskun, Ayse K},
  booktitle={2021 international conference on applied artificial intelligence (ICAPAI)},
  pages={1--8},
  year={2021},
  organization={IEEE}
}

@article{bahri2022shapelet,
  title={Shapelet-based counterfactual explanations for multivariate time series},
  author={Bahri, Omar and Boubrahimi, Soukaina Filali and Hamdi, Shah Muhammad},
  journal={arXiv preprint arXiv:2208.10462},
  year={2022}
}

@article{spinnato2023understanding,
  title={Understanding Any Time Series Classifier with a Subsequence-based Explainer},
  author={Spinnato, Francesco and Guidotti, Riccardo and Monreale, Anna and Nanni, Mirco and Pedreschi, Dino and Giannotti, Fosca},
  journal={ACM Transactions on Knowledge Discovery from Data},
  volume={18},
  number={2},
  pages={1--34},
  year={2023},
  publisher={ACM New York, NY}
}

@inproceedings{li2023attention,
  title={Attention-based counterfactual explanation for multivariate time series},
  author={Li, Peiyu and Bahri, Omar and Boubrahimi, Souka{\"\i}na Filali and Hamdi, Shah Muhammad},
  booktitle={International Conference on Big Data Analytics and Knowledge Discovery},
  pages={287--293},
  year={2023},
  organization={Springer}
}

@inproceedings{wang2017time,
  title={Time series classification from scratch with deep neural networks: A strong baseline},
  author={Wang, Zhiguang and Yan, Weizhong and Oates, Tim},
  booktitle={2017 International joint conference on neural networks (IJCNN)},
  pages={1578--1585},
  year={2017},
  organization={IEEE}
}

@misc{bagnall2018uea,
      title={The UEA multivariate time series classification archive, 2018}, 
      author={Anthony Bagnall and Hoang Anh Dau and Jason Lines and Michael Flynn and James Large and Aaron Bostrom and Paul Southam and Eamonn Keogh},
      year={2018},
      eprint={1811.00075},
      archivePrefix={arXiv},
      primaryClass={cs.LG}
}

@misc{delaney2021instancebased,
      title={Instance-based Counterfactual Explanations for Time Series Classification}, 
      author={Eoin Delaney and Derek Greene and Mark T. Keane},
      year={2021},
      eprint={2009.13211},
      archivePrefix={arXiv},
      primaryClass={cs.LG}
}

@article{das1998normal,
  title={Normal-boundary intersection: A new method for generating the Pareto surface in nonlinear multicriteria optimization problems},
  author={Das, Indraneel and Dennis, John E},
  journal={SIAM journal on optimization},
  volume={8},
  number={3},
  pages={631--657},
  year={1998},
  publisher={SIAM}
}

@INPROCEEDINGS{7966039,
  author={Wang, Zhiguang and Yan, Weizhong and Oates, Tim},
  booktitle={2017 International Joint Conference on Neural Networks (IJCNN)}, 
  title={Time series classification from scratch with deep neural networks: A strong baseline}, 
  year={2017},
  volume={},
  number={},
  pages={1578-1585},
  doi={10.1109/IJCNN.2017.7966039}}

@misc{schäfer2018multivariatetimeseriesclassification,
      title={Multivariate Time Series Classification with WEASEL+MUSE}, 
      author={Patrick Schäfer and Ulf Leser},
      year={2018},
      eprint={1711.11343},
      archivePrefix={arXiv},
      primaryClass={cs.LG},
      url={https://arxiv.org/abs/1711.11343}, 
}

@article{khan2023end,
  title={End-to-end multivariate time series classification via hybrid deep learning architectures},
  author={Khan, Mehak and Wang, Hongzhi and Ngueilbaye, Alladoumbaye and Elfatyany, Aya},
  journal={Personal and Ubiquitous Computing},
  volume={27},
  number={2},
  pages={177--191},
  year={2023},
  publisher={Springer}
}

@article{liu2020dstp,
  title={DSTP-RNN: A dual-stage two-phase attention-based recurrent neural network for long-term and multivariate time series prediction},
  author={Liu, Yeqi and Gong, Chuanyang and Yang, Ling and Chen, Yingyi},
  journal={Expert Systems with Applications},
  volume={143},
  pages={113082},
  year={2020},
  publisher={Elsevier}
}

@inproceedings{zhou2016learning,
  title={Learning deep features for discriminative localization},
  author={Zhou, Bolei and Khosla, Aditya and Lapedriza, Agata and Oliva, Aude and Torralba, Antonio},
  booktitle={Proceedings of the IEEE conference on computer vision and pattern recognition},
  pages={2921--2929},
  year={2016}
}

@article{navas2021optimal,
  title={Optimal counterfactual explanations for scorecard modelling},
  author={Navas-Palencia, Guillermo},
  journal={arXiv preprint arXiv:2104.08619},
  year={2021}
}

@inproceedings{hollig2022tsevo,
  title={Tsevo: Evolutionary counterfactual explanations for time series classification},
  author={H{\"o}llig, Jacqueline and Kulbach, Cedric and Thoma, Steffen},
  booktitle={2022 21st IEEE International Conference on Machine Learning and Applications (ICMLA)},
  pages={29--36},
  year={2022},
  organization={IEEE}
}

@article{deb2002fast,
  title={A fast and elitist multiobjective genetic algorithm: NSGA-II},
  author={Deb, Kalyanmoy and Pratap, Amrit and Agarwal, Sameer and Meyarivan, TAMT},
  journal={IEEE transactions on evolutionary computation},
  volume={6},
  number={2},
  pages={182--197},
  year={2002},
  publisher={Ieee}
}

@article{guidotti2024counterfactual,
  title = {Counterfactual Explanations and How to Find Them: Literature Review and Benchmarking},
  shorttitle = {Counterfactual Explanations and How to Find Them},
  author = {Guidotti, Riccardo},
  year = 2024,
  journal = {Data Mining and Knowledge Discovery},
  shortjournal = {Data Min Knowl Disc},
  volume = {38},
  number = {5},
  pages = {2770--2824},
  publisher = {Springer US},
  issn = {1573-756X},
  doi = {10.1007/s10618-022-00831-6},
  url = {https://link.springer.com/article/10.1007/s10618-022-00831-6},
  urldate = {2025-02-13},
  issue = {5},
  langid = {english}
}

@article{pawelczyk2021carla,
  title={Carla: a python library to benchmark algorithmic recourse and counterfactual explanation algorithms},
  author={Pawelczyk, Martin and Bielawski, Sascha and Heuvel, Johannes van den and Richter, Tobias and Kasneci, Gjergji},
  journal={arXiv preprint arXiv:2108.00783},
  year={2021}
}

@ARTICLE{hamming1950distance,
  author={Hamming, R. W.},
  journal={The Bell System Technical Journal}, 
  title={Error detecting and error correcting codes}, 
  year={1950},
  volume={29},
  number={2},
  pages={147-160},
  doi={10.1002/j.1538-7305.1950.tb00463.x}}

@incollection{wilcoxon1992individual,
  title={Individual comparisons by ranking methods},
  author={Wilcoxon, Frank},
  booktitle={Breakthroughs in statistics: Methodology and distribution},
  pages={196--202},
  year={1992},
  publisher={Springer}
}

@article{friedman1937use,
  title={The use of ranks to avoid the assumption of normality implicit in the analysis of variance},
  author={Friedman, Milton},
  journal={Journal of the american statistical association},
  volume={32},
  number={200},
  pages={675--701},
  year={1937},
  publisher={Taylor \& Francis}
}

@book{nemenyi1963distribution,
  title={Distribution-free multiple comparisons.},
  author={Nemenyi, Peter Bjorn},
  year={1963},
  publisher={Princeton University}
}

@article{zhou2009canonical,
  title={Canonical time warping for alignment of human behavior},
  author={Zhou, Feng and Torre, Fernando},
  journal={Advances in neural information processing systems},
  volume={22},
  year={2009}
}

@article{sakoe2003dynamic,
  title={Dynamic programming algorithm optimization for spoken word recognition},
  author={Sakoe, Hiroaki and Chiba, Seibi},
  journal={IEEE transactions on acoustics, speech, and signal processing},
  volume={26},
  number={1},
  pages={43--49},
  year={2003},
  publisher={IEEE}
}

@inproceedings{cuturi2017soft,
  title={Soft-dtw: a differentiable loss function for time-series},
  author={Cuturi, Marco and Blondel, Mathieu},
  booktitle={International conference on machine learning},
  pages={894--903},
  year={2017},
  organization={PMLR}
}

\end{document}